\documentclass[10pt,letterpaper]{article}
\usepackage[top=0.85in,left=2.75in,footskip=0.75in]{geometry}

\usepackage{amsmath,amssymb}

\usepackage{changepage}

\usepackage[utf8x]{inputenc}

\usepackage{textcomp,marvosym}

\usepackage{cite}

\usepackage{nameref,hyperref}

\usepackage[right]{lineno}

\usepackage{microtype}
\usepackage[table]{xcolor}
\usepackage{array}
\usepackage{multirow}
\usepackage{booktabs, tabularx} % good for beautiful tables

\newcolumntype{+}{!{\vrule width 2pt}}

\newlength\savedwidth

\newcommand\thickhline{\noalign{\global\savedwidth\arrayrulewidth\global\arrayrulewidth 2pt}%
\hline
\noalign{\global\arrayrulewidth\savedwidth}}

\raggedright
\usepackage[aboveskip=1pt,labelfont=bf,labelsep=period,justification=raggedright,singlelinecheck=off]{caption}

\makeatletter
\renewcommand{\@biblabel}[1]{\quad#1.}
\makeatother

\usepackage{lastpage,fancyhdr,graphicx}
\usepackage{epstopdf}
\fancyheadoffset[L]{2.25in}
\fancyfootoffset[L]{2.25in}
\begin{document}
\vspace*{0.2in}

\begin{flushleft}
{\Large
\textbf\newline\textbf\newline{Domain-specific transfer learning in the automated scoring of tumor-stroma ratio from histopathological images of colorectal cancer} % Please use "sentence case" for title and headings (capitalize only the first word in a title (or heading), the first word in a subtitle (or subheading), and any proper nouns).
}
\newline
\\
Liisa Petäinen\textsuperscript{1*},
Juha P. Väyrynen\textsuperscript{2},
Pekka Ruusuvuori\textsuperscript{3,4,5},
Ilkka Pölönen\textsuperscript{1},
Sami Äyrämö\textsuperscript{1},
Teijo Kuopio\textsuperscript{6,7,8}
\\
\bigskip
\textbf{1} Faculty of Information Technology, University of Jyväskylä, Jyväskylä, Finland
\\
\textbf{2} Cancer and Translational Medicine Research Unit, Medical Research Center, Oulu University Hospital, and University of Oulu, Oulu, Finland
\\
\textbf{3} Faculty of Medicine and Health Technology, Tampere University, Tampere, Finland
\\
\textbf{4} Cancer Research Unit, Institute of Biomedicine, University of Turku, Finland
\\
\textbf{5} FICAN West Cancer Centre, Turku University Hospital, Turku, Finland
\\
\textbf{6} Department of Education and Research, Hospital Nova of Central Finland, Jyväskylä,  Finland
\\
\textbf{7} Department of Biological and Environmental Science, University of Jyväskylä, Jyväskylä, Finland
\\
\textbf{8} Department of Pathology, Hospital Nova of Central Finland, Jyväskylä, Finland
\\
\bigskip

% Insert additional author notes using the symbols described below. Insert symbol callouts after author names as necessary.
% 
% Remove or comment out the author notes below if they aren't used.
%
% Primary Equal Contribution Note
%\Yinyang These authors contributed equally to this work.

% Additional Equal Contribution Note
% Also use this double-dagger symbol for special authorship notes, such as senior authorship.
%\ddag These authors also contributed equally to this work.

%\textcurrency Current Address: Dept/Program/Center, Institution Name, City, State, Country % change symbol to "\textcurrency a" if more than one current address note
% \textcurrency b Insert second current address 
% \textcurrency c Insert third current address


% Group/Consortium Author Note
%\textpilcrow Membership list can be found in the Acknowledgments section.

% Use the asterisk to denote corresponding authorship and provide email address in note below.
* lihesalo@jyu.fi

\end{flushleft}
\section*{Abstract}
Tumor-stroma ratio (TSR) is a prognostic factor for many types of solid tumors. In this study, we propose a method for automated estimation of TSR from histopathological images of colorectal cancer. The method is based on convolutional neural networks which were trained to classify colorectal cancer tissue in hematoxylin-eosin stained samples into three classes: \emph{stroma}, \emph{tumor} and \emph{other}. The models were trained using a data set that consists of 1343 whole slide images. Three different training setups were applied with a transfer learning approach using domain-specific data i.e. an external colorectal cancer histopathological data set. The three most accurate models were chosen as a classifier, TSR values were predicted and the results were compared to a visual TSR estimation made by a pathologist. The results suggest that classification accuracy does not improve when domain-specific data are used in the pre-training of the convolutional neural network models in the task at hand. Classification accuracy for \emph{stroma}, \emph{tumor} and \emph{other} reached 96.1\% on an independent test set. Among the three classes the best model gained the highest accuracy (99.3\%) for class \emph{tumor}. When TSR was predicted with the best model, the correlation between the predicted values and values estimated by an experienced pathologist was 0.57. Further research is needed to study associations between computationally predicted TSR values and other clinicopathological factors of colorectal cancer and the overall survival of the patients.

% Please keep the Author Summary between 150 and 200 words
% Use first person. PLOS ONE authors please skip this step. 
% Author Summary not valid for PLOS ONE submissions.   
% \section*{Author summary}
% Lorem ipsum dolor sit amet, consectetur adipiscing elit. Curabitur eget porta erat. Morbi consectetur est vel gravida pretium. Suspendisse ut dui eu ante cursus gravida non sed sem. Nullam sapien tellus, commodo id velit id, eleifend volutpat quam. Phasellus mauris velit, dapibus finibus elementum vel, pulvinar non tellus. Nunc pellentesque pretium diam, quis maximus dolor faucibus id. Nunc convallis sodales ante, ut ullamcorper est egestas vitae. Nam sit amet enim ultrices, ultrices elit pulvinar, volutpat risus.

% \linenumbers

% Use "Eq" instead of "Equation" for equation citations.
\section*{Introduction}

Deep learning (DL) has been the state-of-the-art medical image analysis technology for the last decade. It has been applied to various tasks also in digital pathology, e.g., tissue classification between normal and tumor tissues, defining tumor subtype, recognition (e.g. dividing cells) and segmentation (patch- or pixel-level segmentation)~\cite{litjens2017a}. Tasks other than purely morphological have also been carried out, such as training DL models to predict certain genetic changes from hematoxylin-eosin (H\&E)-stained histopathological sections without e.g. immunohistochemical staining~\cite{chang2018deep, coudray2018classification, schaumberg2018h, schmauch2020deep, wang2020predicting}. These are relevant and meaningful efforts because the aforementioned tasks are time-consuming and expensive when carried out using manual laboratory methods ~\cite{echle2021deep}. 

A common challenge for developing artificial intelligence methods lies in the insatiable data hunger of DL algorithms. The lack of annotated data is also one of the most significant challenges for digital pathology ~\cite{tizhoosh2018artificial}. Gaining a sufficient amount of high-quality training and testing data means hours of work for pathologists to annotate regions of interest to digitized images. One way to alleviate the data scarcity problem could be, e.g., transfer learning.

Transfer learning means utilizing a pre-trained neural network that has already learned a machine learning task in some domain that is not necessarily the same as in the target application. Transfer learning can be accomplished, e.g. using pre-trained ImageNet neural network architecture that will provide the initial parameter values for the model. ImageNet-based transfer learning is a common approach in digital pathology~\cite{russakovsky2015a, bayramoglu2016transfer, kieffer2017convolutional}. Although ImageNet model has been trained with images representing dogs, planes and houses, that are essentially very dissimilar to  histopathological images, initializing weights of the convolutional neural network (CNN) model with ImageNet has been shown to increase prediction accuracy in medical imaging tasks ~\cite{litjens2017a, mormont2018comparison}. Sometimes a pre-trained model can also be available from the target domain. Some studies have shown that such a domain-specific pre-trained model may further improve prediction performance in the context of histopathological image analysis when compared to the ImageNet-initialization~\cite{mormont2018comparison, sharmay2021histotransfer}.

This study focuses on automating the estimation of TSR from histopathological images of colorectal cancer (CRC) using transfer learning. CRC is the second most death-causing cancer in the world with over 900,000 deaths every year~\cite{organization2020a}. One prognostic factor for CRC is the proportion of stroma within the tumor site. It has been shown to associate with the survival of the patient in many solid cancer types. The low amount of stroma (TSR $\leq$ 50 \%) associates with a better prognosis~\cite{west2010proportion, huijbers2013a, ma2012a, mesker2007a}.

Pathologists determine TSR visually by following a certain scoring protocol~\cite{van2018a}. The main problem of this approach is the reproducibility of the TSR scoring. Overview by Van Pelt et al.~\cite{van2018a} showed that the Cohen's kappa scores measuring the inter-observer agreement of visual TSR using binary scoring (TSR $>$ 50 \% = stroma-high and TSR $\leq$ 50 \% = stroma-low) ranged from 0.60 to 0.89. Automated TSR estimation may improve reproducibility, but it is a challenging task and a  new relatively new concept. 

Automated estimation of TSR begins by tiling a histopathological whole-slide image (WSI) into smaller image patches. After that TSR can be predicted by classifying the patches and calculating the proportion of tumor and stroma. Another approach is to use a particular spot that is a smaller part of the WSI selected by a pathologist to calculate TSR. Both approaches have been applied in previous studies~\cite{sirinukunwattana2015spatially, sirinukunwattana2018novel,kather2019predicting, zhao2020a, park2014a, geessink2019computer}.

Sirinukuvattana et al.~\cite{sirinukunwattana2015spatially, sirinukunwattana2018novel} automated the TSR estimation using a CNN model trained for nuclei detection. They trained a model based on VGG19~\cite{simonyan2014very} architecture to classify nine tissue types and the accuracies for detecting stroma and tumor were 90.4 \% and 96.0 \%, respectively. In contrast to other TSR-related studies, their results did not show prognostic value for TSR~\cite{kather2019predicting, zhao2020a, park2014a}. 

Zhao et al.~\cite{zhao2020a} proposed a nine-class CNN model for which overall classification accuracies on two test sets were 95.7 \% and 97.5 \%. Classification accuracies for tumor and stroma were 92.8 \% and 70.9 \% on the test set 1 that was published by Kather et al.~\cite{kather2018a}. The test set 2 was a random sample from images collected at Yunnan Cancer Hospital. Classification accuracies for  tumor and stroma on the test set 2 were 97.2 \% and 89.1 \%. Zhao et al. used pathologists annotations as ground truth for TSR estimation on 126 image blocks of size 1 $\mu m^2$, the predicted tumor and stroma areas were in high agreement with pathologists’ annotations (Pearson r = 0.939, 95\% CI 0.914 - 0.957). When splitting the patient cohort into two categories stroma-low (TSR < 48.8\%) and stroma-high (TSR$\ge$48.8\%) based on the TSR results of their model, TSR was shown to be an independent prognostic factor in the overall survival of CRC patient.

Instead of using the whole slide as an input for a neural network model, an alternative approach is to calculate TSR from the same circular spot where the visual TSR estimation takes place. The downside of this approach is the manual effort needed to point the spot for the machine learning model. Geessink et al.~\cite{geessink2019computer} developed an 11-layer VGG neural network on 129 patients to classify nine tissue types. Using 50 \% cutoff-value between stroma-high and stroma-low categories there was a considerable disagreement between the model and pathologist (Cohen’s kappa $\kappa$ = 0.239). Cohen's kappa score was slightly improved ($\kappa$ = 0.521) using the median of the TSR values estimated with the model as a cutoff value for stroma-high and stroma-low. Also, stroma-high- and stroma-low grouping showed a strong prognostic value when the median was used as a cutoff value.

In the present study, 12 DL models for estimating TSR for CRC samples were developed and tested using three different transfer learning setups, four different pre-trained CNN architectures, and two distinct CRC data sets. Models were trained to classify image patches into three different tissue classes: \emph{tumor}, \emph{stroma} and \emph{other}. Three best-performing CNN models were chosen to estimate TSR from WSIs in a separate TSR test set. The predicted TSR values were then compared with the pathologist's TSR estimates.

\section*{Materials and methods}

\subsection*{Data}

% Data on monikko englannissa
% Data set kirjoitetaan yleisemmin erikseen, muutenkin vuoksi suosisin pelkkää data sanaa
% Virkettä ei pitäisi aloittaa numeroilla
% tile ja patch
% For figure citations, please use "Fig" instead of "Figure".
In this study, two mutually independent CRC data sets were used: a CRC data set from Central Finland Health Care District ("CFHCD-data") and a public CRC data set ("NCT-CRC-HE-100K")~\cite{kather2018a}.

CFHCD-data include 1343 anonymized WSIs, resolution 0.5 microns per pixel (MPP), of primary colorectal cancers (stages I-IV). The cohort is described in detail by Elomaa et al.~\cite{elomaa2022prognostic}. The WSIs were scanned with Hamamatsu NanoZoomer-XR. 

Automated estimation of the TSR-value for a single WSI is based on calculating the ratio of patches representing stroma and tumor classes. The estimation process consists of two steps: 1) CNN-classification of patches for a WSI 2) Calculation of the ratio of tumor and stroma patches.

\begin{figure}[!h]
\includegraphics[height=9cm,keepaspectratio]{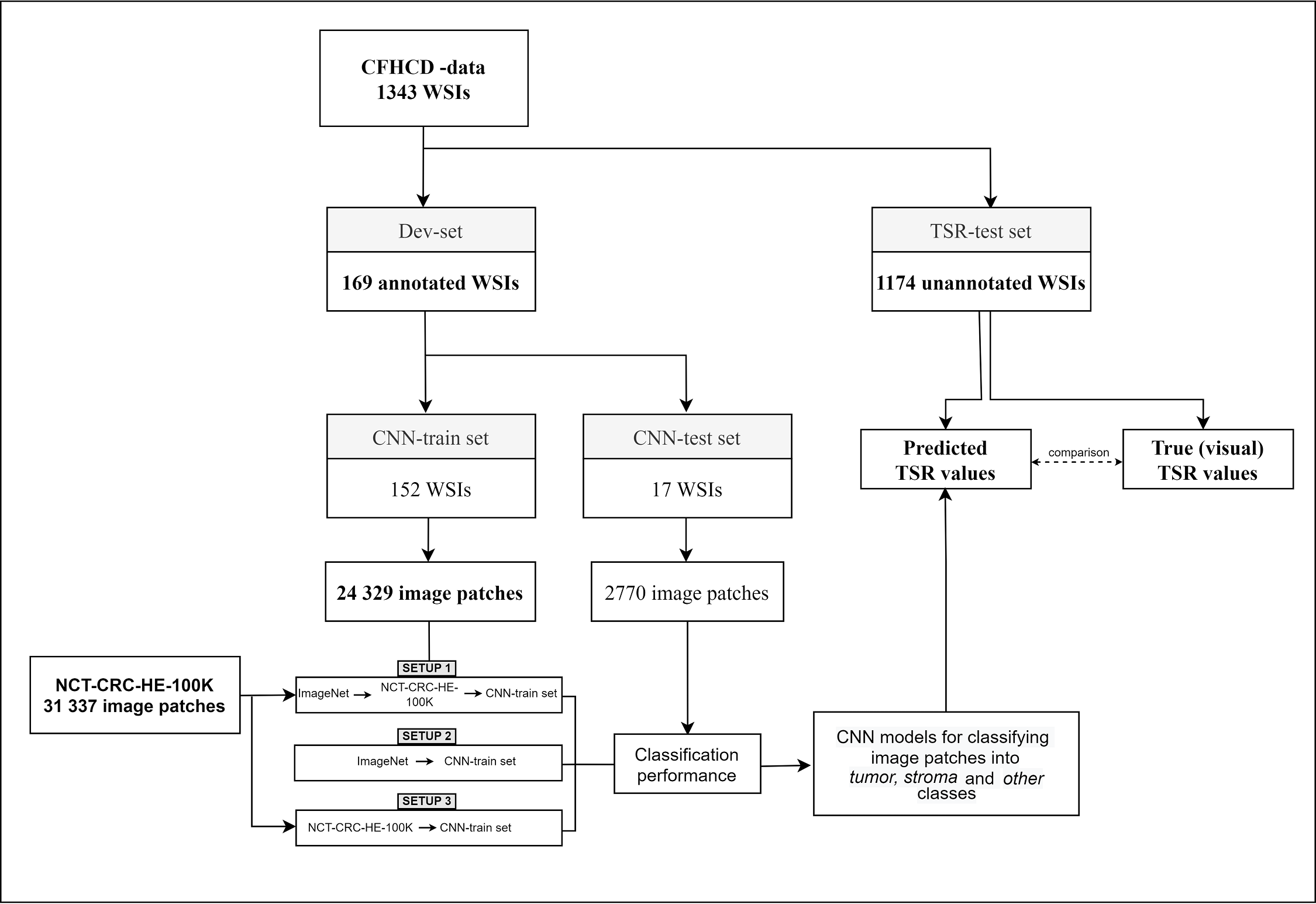}
\caption{{\bf Diagram of the studyflow.}
Patches for training were tiled from 152 annotated WSIs (CNN-train), the classifier test set of 17 WSIs (CNN-test) was excluded from the training data. TSR-values were predicted for the remaining 1174 WSIs (TSR-test) and compared with the TSR-values determined visually by a pathologist.}
\label{study_flow}
\end{figure}

For developing and testing the models, CFHCD-data were split into two disjoint sets at random. The first one (Dev-set) consisted of 169 WSIs and was used in developing CNN classification models and the second one (TSR-test set) consisted of 1174 WSIs and it was applied to test CNN-based estimation of the TSR-value. The overall study flow is presented in Fig~~\ref{study_flow}.

% Place figure captions after the first paragraph in which they are cited.

\begin{figure}[!h]
\includegraphics[height=5.2cm,keepaspectratio]{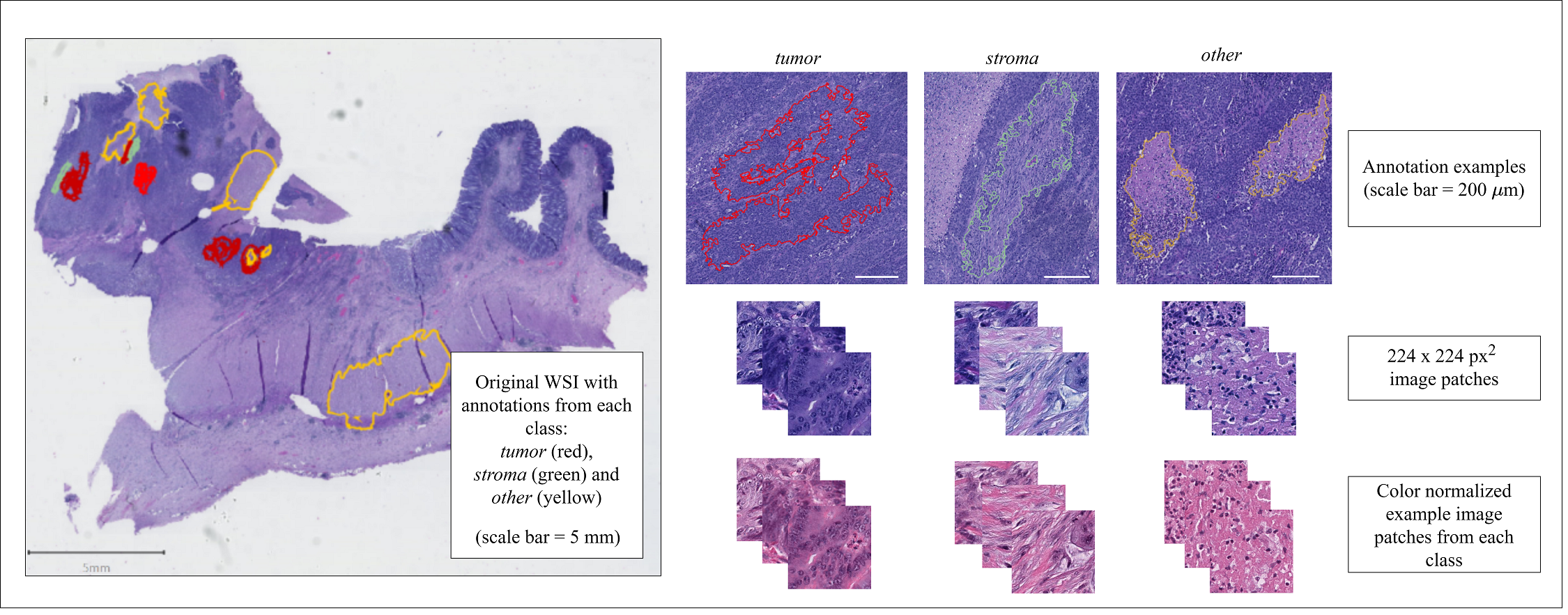}
\caption{{\bf Annotation examples}
Examples of annotations from each class. The patches were extracted from annotated areas, and they were color normalized using Macenko's method~\cite{macenko2009method}.}
\label{annotation_example}
\end{figure}

Before tiling and preprocessing, Dev-set was annotated by an experienced pathologist into three different tissue categories, \emph{tumor}, \emph{stroma} and \emph{other}. The annotations were accomplished with QuPath image analysis tool~\cite{bankhead2017qupath}. The class \emph{other} includes debris, lymphocytes, mucus, normal epithelium and smooth muscle (See Fig ~~\ref{annotation_example}). Ground truth TSR-values (from now on referred to as true TSR-values) were visually assessed by the pathologist for the TSR-test set following the protocol by Van Pelt et al.~\cite{van2018a}.

A subsample (n = 31,337) of NCT-CRC-HE-100K data set were used to investigate the effects of domain-specific pre-training of the CNN models. The original data includes 100,000 patches tiled from 86 WSIs (0.5 MPP). The patches are pre-labeled to nine tissue classes: adipose, debris, lymphocytes, mucus, normal, smooth muscle, stroma and tumor. In this study, only patches representing \emph{stroma} (10,446 patches) and \emph{tumor} (10,446 patches) classes were used as such, whereas a random sample of 10,445 patches from debris, lymphocytes, mucus, normal and smooth muscle classes with even distribution was drawn and assigned to a class called \emph{other}.

\subsection*{Tiling and preprocessing of WSIs}
% Tässä kuvaus mitä WSI:lle tapahtuu että siitä tulee input neuroverkolle - jos käsittelyssä eroja CFHCD ja Kahter niin mainitse myös erot
Following the annotation of the WSIs in Dev-set they were tiled into smaller WSI patches (224 x 224 pixels / 101 $\mu m$ x 101 $\mu m$). Tiling was performed with a sliding window procedure with 64 pixels overlapping. If less than 75\% of the patch area included annotated pixels, the patch was discarded from further analyses. All patches were color normalized by Macenko's method~\cite{macenko2009method}.

Since the classifiers were not trained to detect the image background and adipose tissue, the WSIs in the TSR-test set were tiled in the following way: First, to remove image background and adipose tissue, a binary mask was applied using Otsu's algorithm for choosing the optimal threshold value~\cite{otsu1979threshold}. After this the image patches were tiled using a sliding window procedure with no overlapping. If the masked area was less than 75\%, the patch was discarded from further analyses. Macenko's method was applied for color normalization of the patches~\cite{macenko2009method}.

\subsection*{Training and evaluation of classifiers}

For training and selecting the best CNN models Dev-set was split at random into two distinct WSI sets of CNN-train (n=152) and CNN-test (n=17) with the constraints that after the tiling process the maximum difference of eighty patches in the training distribution of the target classes was allowed and the minimum number of patches was nine hundred in each test class. The constraints were applied to ensure approximately balanced class distribution in CNN-train and that the size of CNN-test is at least 10\% of Dev-set. 

\begin{figure}[!h]
\includegraphics[height=4.2cm,keepaspectratio]{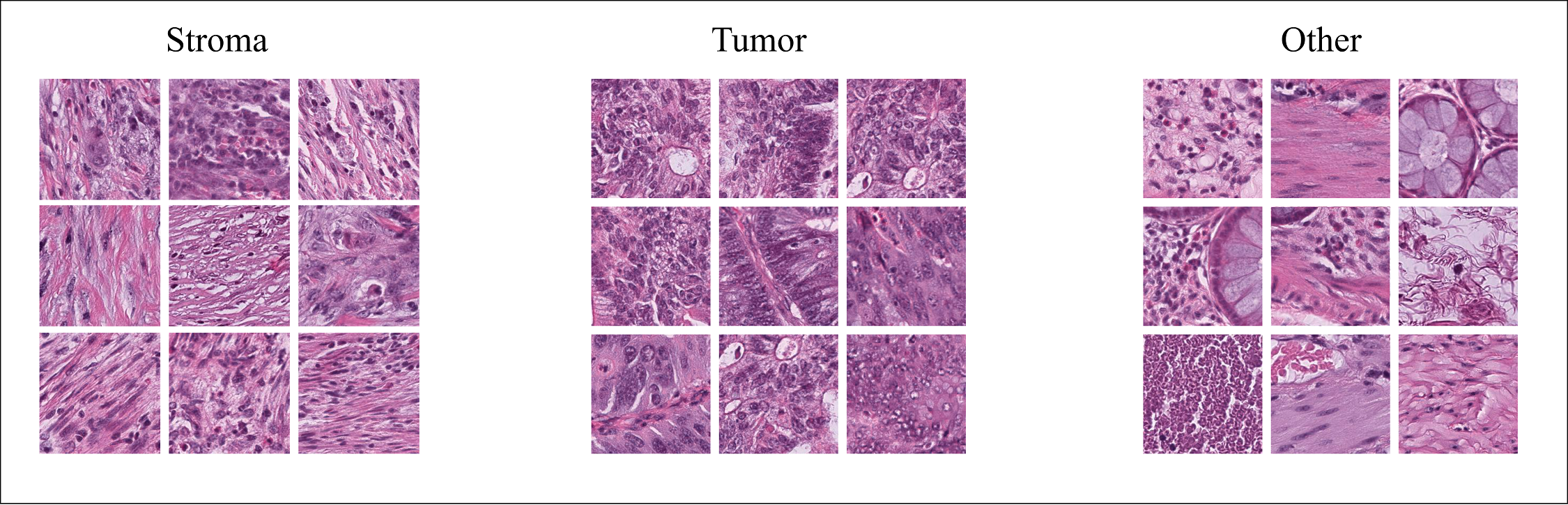}
  \caption{{\bf Example patches from each class}
The image patches were taken from annotated areas and the number of patches was balanced between the groups. The \emph{other} class includes, e.g., debris, normal epithelium and smooth muscle.}
\label{exampletiles}
\end{figure}

This resulted in 24,329 and 2,770 input patches for training and testing of the classifiers, respectively. The final numbers of patches in the CNN-train/CNN-test classes were 8139/900, 8060/900, and 8130/900 for \emph{tumor}, \emph{stroma} and \emph{other}, respectively. Examples of image patches from each class are shown in ~~Fig\ref{exampletiles}.

% Place figure captions after the first paragraph in which they are cited.

For predicting TSR in WSIs, four convolutional neural network (CNN) architectures, AlexNet~\cite{krizhevsky2012imagenet}, GoogleNet~\cite{szegedy2015going}, ResNet50~\cite{he2016deep}, VGG19~\cite{simonyan2014very}, were trained on CNN-train to classify the patches into three tissue types: \emph{tumor}, \emph{stroma} and \emph{other}. For all the CNN models, the input layer was set according to the WSI patch dimensions $224 \times 224$, and the output layer was softmax function (dimension=3).

Three different pre-training strategies for CNN models were applied (SETUP-1,  SETUP-2 and SETUP-3). In SETUP-1 the CNN model was first initialized with ImageNet~\cite{russakovsky2015imagenet} weights and thereafter pre-trained with the domain-specific subsample of NCT-CRC-HE-100K-data before finalizing the training with CNN-train. In SETUP-2 training on CNN-train started directly from ImageNet weights. In SETUP-3 the CNNs were first initialized with random weights using PyTorch default parameters and then subsequently pre-trained with the subsample of NCT-CRC-HE-100K-data and finalized with CNN-train.

%The model development were accomplished using the training set of 31,337 image patches from Kather-data and 24,329 image patches from CFHCD-data. 

For both pretraining the CNN models on the NCT-CRC-HE-100K patches as well as fine-tuning the models on CFHCD-data (CNN-train and CNN-test), the most suitable values for hyperparameters were selected using 5-fold cross-validation (C-V) CNN-train. In 5-fold C-V data are split at random into the five subsets. Then each subset acts once as a test fold (set) while the four other are used for training. The error estimate for a model is the average error over the test folds. 

After finding the best hyperparameter values for the neural networks, the optimal number of epochs was determined with an early stopping model selection strategy on CNN-train using 1/3 of the training patches for validation. The final model was then trained once more on the full CNN-train until the optimal number of epochs was reached.

The generalization performance of each CNN model was then assessed by computing the classification accuracy, precision, recall and  $F_{1}$-score of the models on CNN-test (2,770 CFHCD-patches).

For more information about chosen hyperparameters, see \nameref{S1_Table}. 

\subsection*{Calculating tumor-stroma ratio}

TSR was calculated for each WSI in TSR-test as the ratio of patches classified by a CNN model as \emph{stroma} and \emph{tumor}:

\begin{eqnarray}
\centerline{ $TSR = \frac{n_{stroma}}{n_{tumor} + n_{stroma}}$},
\end{eqnarray}

where $n_{stroma}$ and $n_{tumor}$ are the number of \emph{stroma} and \emph{tumor} patches, respectively.

\subsection*{Equipment and software}
 All the neural network models were trained on Linux GPU server Tesla P100, x 86\_64 with Python-version 3.8.5. using PyTorch- and TorchVision-libraries, versions 1.9.0 and 0.10.0, respectively. OpenCV 4.5.2 was utilized when masking the WSIs. Color normalization was applied with an open-source library StainTools, available for download at GitHub: https://github.com/Peter554/StainTools. Performance metrics were calculated with Scikit-learn 0.24.2 metrics-module.

% Place figure captions after the first paragraph in which they are cited.
%\begin{figure}[!h]
%\caption{{\bf Bold the figure title.}
%Figure caption text here, please use this space for the figure panel descriptions instead of using %subfigure commands. A: Lorem ipsum dolor sit amet. B: Consectetur adipiscing elit.}
%\label{fig1}
%\end{figure}

% Results and Discussion can be combined.
\section*{Results}

\subsection*{Validation and test results of the final models}

Classification accuracy metrics on the CNN-test set are shown for all the final CNN models in Table~~\ref{test}. The three highest values were obtained by training the models directly from Imagenet-pretrained weights. The differences between the three best CNN architectures are negligible. Only Alexnet showed slightly poorer performance in terms of classification accuracy. To support the interpretation of the results also validation accuracies are reported in Table~~\ref{val}.

\begin{table}[!ht]
%\begin{adjustwidth}{-2.25in}{0in} % Comment out/remove adjustwidth environment if table fits in text column.
\centering
\caption{
{\bf Test accuracies}}
\begin{tabular}{|l|l|l|l|}
\hline
& \multicolumn{1}{l|}{\bf SETUP 1} & \multicolumn{1}{l|}{\bf SETUP 2} & \multicolumn{1}{l|}{\bf SETUP 3}\\
& \multicolumn{1}{l|}{accuracy} & \multicolumn{1}{l|}{accuracy} & \multicolumn{1}{l|}{accuracy} \\ \thickhline
    Alexnet		& 91.95 \% & 92.42 \% & 91.79 \%  \\ \hline
    Googlenet   & 94.94 \% & \textbf{95.40} \% & 95.37 \% \\ \hline
    ResNet50	& 94.94 \% & \textbf{96.09} \% & 92.57 \% \\ \hline
    VGG\-19		& 95.19 \% & \textbf{95.65} \% & 92.64 \% \\ \hline
\end{tabular}
\begin{flushleft} Test accuracies of all final models on the CNN-test set, top-3 models are shown bold.
\end{flushleft}
\label{test}
%\end{adjustwidth}
\end{table}

\begin{table}[!ht]
%\begin{adjustwidth}{-2.25in}{0in} % Comment out/remove adjustwidth environment if table fits in text column.
\centering
\caption{
{\bf Validation accuracies}}
\begin{tabular}{|l|l|l|l|}
\hline
& \multicolumn{1}{l|}{\bf SETUP 1} & \multicolumn{1}{l|}{\bf SETUP 2} & \multicolumn{1}{l|}{\bf SETUP 3}\\
& \multicolumn{1}{l|}{accuracy} & \multicolumn{1}{l|}{accuracy} & \multicolumn{1}{l|}{accuracy} \\ \thickhline
    Alexnet		& 90.19 \% & 92.90 \% & 92.30 \% \\ \hline
    Googlenet	& 93.35 \% & 92.19 \% & \textbf{93.69} \% \\ \hline
    ResNet50	& \textbf{93.39} \% & 92.96 \% & 91.49 \% \\ \hline
    VGG\-19		& \textbf{94.17} \% & 93.29 \% & 91.11 \% \\ \hline
\end{tabular}
\begin{flushleft} Validation accuracies for all final CNN classification models (top-3 models are bolded).
\end{flushleft}
\label{val}
%\end{adjustwidth}
\end{table}

\begin{figure}[!h]
\includegraphics[height=4.98cm,keepaspectratio]{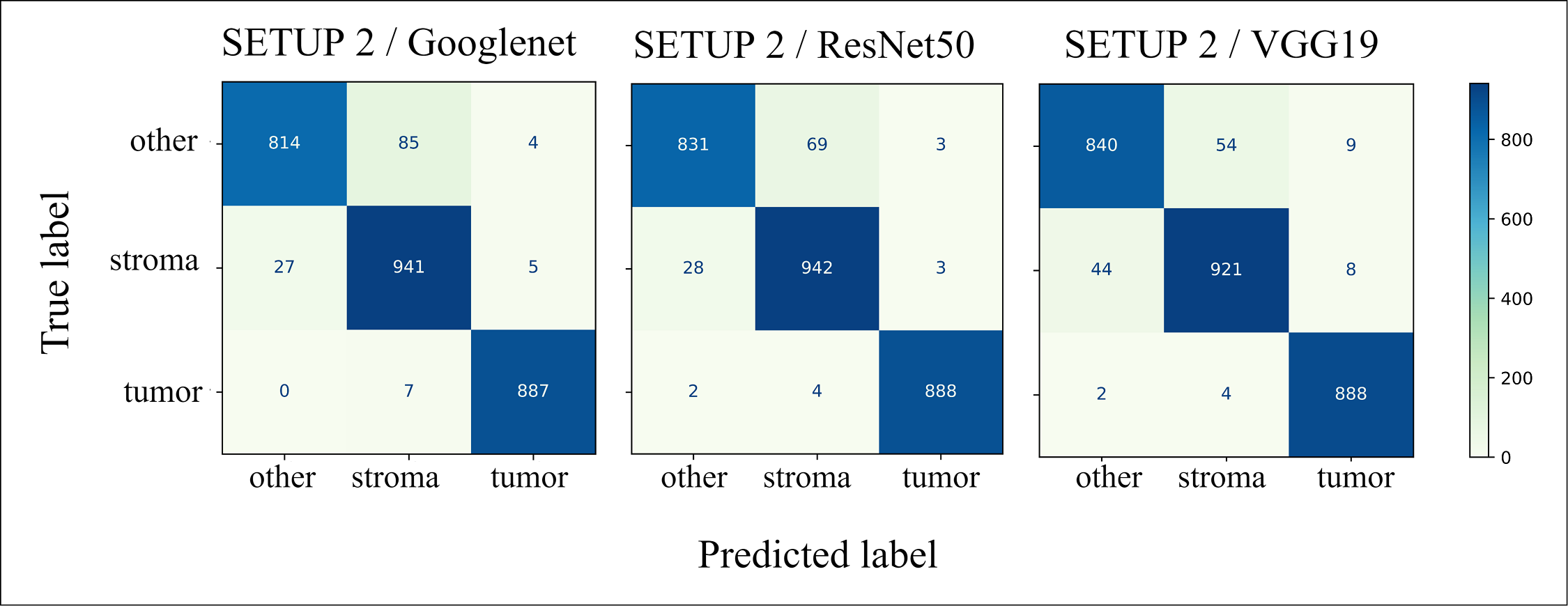}
  \caption{{\bf Confusion matrices.}
Classification results of top-3 models on the CNN-test set.}
\label{confusion_matrices}
\end{figure}

A more detailed comparison between true and predicted classifications on CNN-test can be seen for the most accurate top-3 models in Fig~~\ref{confusion_matrices}. All the models performed well in detecting the \emph{tumor} class but had slight difficulties distinguishing between classes \emph{stroma} and \emph{other}.

% Place figure captions after the first paragraph in which they are cited.

Precision, recall and $F_{1}$-score are shown in Table~~\ref{precision_recall_F1}. The numbers show that all the top-3 models attained excellent hit rate (recall > 0.9) for all classes as well as they are accurate in predicting positive cases (precision > 0.9). Consequently all the models attained high F1-score (> 0.9). The observed differences between the top-3 models are so small that they can most likely be explained by random variation.

\begin{table}[!ht]
\caption{
{\bf Precision, recall and $F_{1}$-score}}
%\begin{adjustwidth}{-2.25in}{0in} % Comment out/remove adjustwidth environment if table fits in text column.
\centering
\begin{tabular}{|c|r|r|r|r|}
\hline
%\multicolumn{5}{|c|}{\bf{SETUP 2 / Googlenet}}\\ \hline
    %\multicolumn{2}{|l}{}\multicolumn{10}{l|}{Visual TSR}\\
    %\hline
    \bf{model} & \bf{class} & \bf{precision} &  \bf{recall} &  \bf{$F_{1}$-score} \\ 
    \thickhline
    \multirow{3}{5em}{SETUP 2 / Googlenet}
    & other & 0.97 & 0.90 & 0.93 \\ \cline{2-5}
    & stroma & 0.91 & 0.97 & 0.94 \\ \cline{2-5}
    & tumor & 0.99 & 0.99 & 0.99 \\ \cline{2-5} \hline
    %\hline
    % \multirow{3}{6em}{SETUP 2 / ResNet50} & & & & \\ 
    % \bf{precision} &  \bf{recall} & \bf{f1-score}\\
    \cline{2-5} % \hline
    \multirow{3}{5em}{SETUP 2 / ResNet50} & other & 0.97 & 0.92 & 0.94 \\ \cline{2-5} %\hline
    & stroma & 0.93 & 0.97 & 0.95 \\ \cline{2-5} %\hline
    & tumor & 0.99 & 0.99 & 0.99 \\ \cline{2-5} \hline
    % \multirow{3}{6em}{SETUP 2 / VGG19} & & & & \\
    % \bf{precision} &  \bf{recall} &  \bf{f1-score}\\ 
    \cline{2-5}
    \multirow{3}{5em}{SETUP 2 / VGG19} & other & 0.95 & 0.93 & 0.94 \\ \cline{2-5}
    & stroma & 0.94 & 0.95 & 0.94 \\ \cline{2-5}
    & tumor & 0.98 & 0.99 & 0.99 \\ \cline{2-5}
    \hline
  \end{tabular}
\begin{flushleft} Precision, recall and $F_{1}$-score on the CNN-test set of top-3 models. The CNN-test set included 2770 image tiles.
\end{flushleft}
\label{precision_recall_F1}
%\end{adjustwidth}
\end{table}

\subsection*{Tumor-stroma ratio predictions}

\begin{figure}[!h]
\includegraphics[height=3.9cm,keepaspectratio]{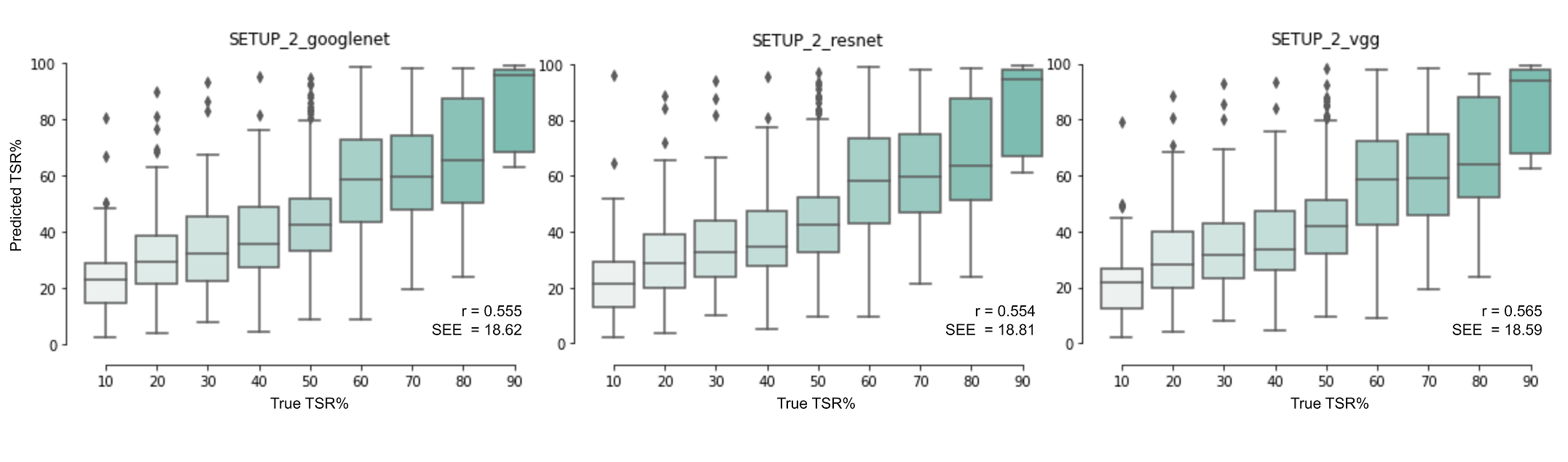}
\caption{{\bf Results from TSR prediction: boxplots}
Boxplots showing the distributions of the predicted TSR values and Pearson correlation coefficient (r) of predicted and true TSR values for the top-3 models on the TSR-test set. The results are shown for each visually estimated TSR value category on the discrete scale $\{10\%, 20\% , \ldots, 90 \%\}$. The standard error of the estimate (SEE) is the overall SEE of all categories.}
\label{boxplots}
\end{figure}

Distributions of the predicted TSR values on the TSR-test set are shown for the top-3 models in Fig~~\ref{boxplots}. Examples of the most accurate and least accurate TSR predictions on WSIs shown in \nameref{S2_Figure}.

Mean, median, standard error of the estimate (SEE), standard deviation (std) of predicted TSR values and the number of WSIs in each category are shown in Table~~\ref{stats_top3}. The results show that all the models overestimate the TSR value in the three lowest target categories (10\%, 20\% and 30\%) and underestimate in the remaining target categories. All the models perform best in the category 60\% (the mean differences between the predicted and true TSR values 1.4) whereas the poorest performance is observed in the categories 10\% and 80\% (the mean differences between the predicted and true TSR values 14.4 and 14.2 respectively). The smallest SEEs are observed in categories 30\%, 40 \% and 90 \%. When grouping the TSR values into stroma-high (TSR $>$ 50 \%) and stroma-low (TSR $\leq$ 50 \%), the Cohen's kappa scores between the true and predicted TSR values were 0.32 (SETUP 2 / Googlenet), 0.33 (SETUP 2 / ResNet50) and 0.33 (SETUP 2 / VGG19).

\begin{table}[!ht]
% \begin{adjustwidth}{-2.25in}{0in} % Comment out/remove adjustwidth environment if table fits in text column.
\centering
\caption{
{\bf TSR results: statistics.}}
\begin{tabular}{|l|r|r|r|r|r|r|r|r|r|r|}
\hline
\multicolumn{11}{|l|}{\bf{SETUP 2 / Googlenet}}\\ \hline
    %\multicolumn{2}{|l}{}\multicolumn{10}{l|}{Visual TSR}\\
    %\hline
    \multicolumn{2}{|l|}{\bf{True TSR}} & \bf{10\%} &  \bf{20\%} &  \bf{30\%} &  \bf{40\%} &  \bf{50\%} &  \bf{60\%} &  \bf{70\%} &  \bf{80\%} &  \bf{90\%} \\ 
    \hline
    \multirow{5}{4em}{Predicted TSR} &  mean & 24.6 & 32.0 & 35.4 & 38.4 & 44.1 & 58.5 & 60.5 & 65.8 & 86.4 \\ \cline{2-11}
    &  median & 23.0 & 28.9 & 32.4 & 35.9 & 42.2 & 58.6 & 59.6 & 65.4 & 96.0 \\ \cline{2-11}
    &  SEE & 20.7 & 20.5 & 16.8 & 15.1 & 17.2 & 19.3 & 21.3 & 25.0 & 14.7 \\ \cline{2-11}
    &  std & 14.9 & 16.6 & 16.0 & 15.1 & 16.1 & 19.3 & 19.1 & 20.8 & 14.8 \\ \cline{2-11}
    &  n & 53 & 95 & 105 & 200 & 284 & 229 & 143 & 51 & 13 \\ \hline  
    \multicolumn{11}{|l|}{\bf{SETUP 2 / ResNet50}}\\ \hline
    \multicolumn{2}{|l|}{\bf{True TSR}} & \bf{10\%} &  \bf{20\%} &  \bf{30\%} &  \bf{40\%} &  \bf{50\%} &  \bf{60\%} &  \bf{70\%} &  \bf{80\%} &  \bf{90\%} \\ 
    \hline
    \multirow{5}{4em}{Predicted TSR} & mean & 24.7 & 31.3 & 35.5 & 38.5 & 44.4 & 58.6 & 60.6 & 66.0 & 86.0 \\ \cline{2-11}
    & median & 21.7 & 28.3 & 32.4 & 34.8 & 42.3 & 58.8 & 59.6 & 63.6 & 94.7 \\ \cline{2-11}
    &  SEE & 21.9 & 20.2 & 17.0 & 15.4 & 17.4 & 19.4 & 21.4 & 24.8 & 15.5 \\ \cline{2-11}
    & std & 16.3 & 16.8 & 16.2 & 15.3 & 16.5 & 19.4 & 19.4 & 20.7 & 15.6 \\ \cline{2-11}
    & n & 53 & 95 & 105 & 200 & 284 & 229 & 143 & 51 & 13 \\ \hline  
    \multicolumn{11}{|l|}{\bf{SETUP 2 / VGG19}}\\ \hline
    \multicolumn{2}{|l|}{\bf{True TSR}} & \bf{10\%} &  \bf{20\%} &  \bf{30\%} &  \bf{40\%} &  \bf{50\%} &  \bf{60\%} &  \bf{70\%} &  \bf{80\%} &  \bf{90\%} \\
    \hline
    \multirow{5}{4em}{Predicted TSR} &  mean & 23.3 & 31.0 & 34.4 & 37.1 & 43.1 & 57.7 & 60.1 & 65.9 & 85.7 \\ \cline{2-11}
    &  median & 22.0 & 28.3 & 31.6 & 33.6 & 41.7 & 58.8 & 58.9 & 64.1 & 94.0 \\ \cline{2-11}
    &  SEE & 19.0 & 19.6 & 16.5 & 15.1 & 17.4 & 19.4 & 21.7 & 25.1 & 14.8 \\ \cline{2-11}
    &  std & 13.7 & 16.3 & 16.0 & 14.9 & 16.0 & 19.3 & 19.4 & 21.0 & 14.8 \\ \cline{2-11}
    &  n & 53 & 95 & 105 & 200 & 284 & 229 & 143 & 51 & 13 \\   
    \hline
  \end{tabular}
\begin{flushleft} Main statistics from top-3 models on the TSR-test set. The results are shown based on the visually estimated TSR-value on the discrete scale $\{10\%, 20\% , \ldots, 90 \%\}$. The overall SEE of all categories combined shown in Fig~~\ref{boxplots}.
\end{flushleft}
\label{stats_top3}
%\end{adjustwidth}
\end{table}
%\subsection*{Computing times}

%Table~~\ref{computing_times} lists average training times for one epoch and average predicting times for one image tile of top-3 CNN-models. 

%\begin{table}[ht]
%\centering
%\caption{
%{\bf Computing times.}}
%\begin{tabular}{|l|c|c|}
%\hline
%& \multicolumn{2}{l|}{\bf{SETUP 2}} \\ 
%%\cline{2-7}
%& Train / & Predict /  \\ 
%& epoch (s) & tile (ms) \\ \thickhline
%Googlenet & 114 & 2.39 \\ \hline
%Resnet50 & 332 & 1.95 \\ \hline
%VGG19 & 450 & 3.01 \\ \hline
%\end{tabular}
%\begin{flushleft} The average training time per epoch and the average predicting time per image tile in milliseconds of top-3 models.
%\end{flushleft}
%\label{computing_times}
%\end{table}

\section*{Discussion}

The aim of this study was to investigate how accurately CNN-based machine learning models can predict the ratio of tumor and stroma tissue in WSI samples. For the best model (ResNet50 architecture) the correlation between the true and predicted TSR values was 0.57 (SEE = 18.6). Approximately the same performance was obtained with GoogleNet and VGG19 architectures. Utility of domain-specific pre-training of CNN models was also investigated, but no meaningful differences were observed. The results show that the same or even better performance with comparable computational cost can be achieved in the present task without domain-specific pre-training of CNN.

Even though the automated TSR were predicted from the whole tissue area in contrast to the pathologist who selected a small spot for estimation, their outcomes correlates ($r=0.57$) rather well. Cohen's kappa score ($\kappa$ = 0.33) for stroma-high and stroma-low classification was comparable with results from the previous study by Geessink et al.~\cite{geessink2019computer} where Cohen's kappa score for stroma-high and stroma-low with 50 \% cutoff-value was 0.24. An overview by~\cite{van2018a} show that the inter-observer kappa-values variate between 0.60 to 0.89 when pathologists estimate the TSR of CRC binary into stroma-high and stroma-low.

In this study SEE was lowest in TSR categories 30\%, 40\% and 90\%. When comparing the means of the predicted TSR values with the true TSR values, the best performing set of images was the one with TSR 60\% and the second largest number of samples. Balancing the data for TSR prediction task might bring more consistency to the results.

The results also indicate that the most difficult part of the classification task is to separate \emph{stroma} and \emph{other}. This can be due to the visual similarity of smooth muscle and fibrotic stroma. This may cause the classifier to make a mistake since the smooth muscle tissue belongs to the \emph{other} class. The smooth muscle and stroma are difficult to separate even by the human eye. Despite the minor weaknesses, the classification accuracy of the best CNN models was comparable to previous studies~\cite{kather2019predicting, zhao2020a}. The classification accuracy for the \emph{tumor} tissue was over 98\% with all top-3 models.

Despite the promising performance of machine learning bringing some aspects of the visual estimation procedure could improve the automated models. For example, only tumor-related stroma tiles could be taken into account. Another option could be mimicking the visual human process by going through the image frame by frame and choosing one spot for making the final TSR prediction. In addition to these, using a smaller tile size might increase classifying accuracy since some tumor areas, as well as stromal areas in between, seem to be quite narrow. This could have a significant effect on the quality of TSR predictions.

Even though accuracy of the proposed automated method for TSR estimation does not fully compare to human visual analysis, the reproducibility of computational model outcomes is a major advantage. Automated machine learning based tools would bring reproducibility to daily practices and the TSR estimation process, in particular. Moreover, TSR estimation with an automated machine learning model can be completed in a fraction of the time compared to the visual method.

As visually estimated TSR has been shown to be an independent prognostic factor in solid cancer types~\cite{west2010proportion, huijbers2013a, ma2012a, mesker2007a}, further studies should take place for assessing the correlation of the machine learning predicted TSR values with other clinicopathological factors and the overall survival of patients.

\section*{Supporting information}

% Include only the SI item label in the paragraph heading. Use the \nameref{label} command to cite SI items in the text.

\paragraph*{S1 Table.}
\label{S1_Table}
{\bf Hyperparameters.}

\paragraph*{S2 Figure.}
\label{S2_Figure}
{\bf Examples of tumor-stroma ratio prediction.}

\section*{Acknowledgments}

This study is one part of AI Hub Central Finland project that has received funding from the Council of Tampere Region (Decision number: A75000) and Leverage from the EU 2014–2020, funded by European Regional Development Fund (ERDF). 

We thank Central Finland Health Care District and "Suolisyöpä Keski-Suomessa 2000-2015" -project for providing the data set. 

\nolinenumbers

% Either type in your references using
% \begin{thebibliography}{}
% Text
% \end{thebibliography}
%
% or
%
% Compile your BiBTeX database using our plos2015.bst
% style file and paste the contents of your .bbl file
% here. See http://journals.plos.org/plosone/s/latex for 

\bibliographystyle{plos2015} % We choose the "plain" reference style
\bibliography{main}

% \begin{refs}
% \end{refs}
% \begin{thebibliography}{10}

% \bibitem{bib1}
% Conant GC, Wolfe KH.
% \newblock {{T}urning a hobby into a job: how duplicated genes find new
%  functions}.
%\newblock Nat Rev Genet. 2008 Dec;9(12):938--950.

%\bibitem{bib2}
%Ohno S.
%\newblock Evolution by gene duplication.
%\newblock London: George Alien \& Unwin Ltd. Berlin, Heidelberg and New %York:
%  Springer-Verlag.; 1970.

%\bibitem{bib3}
%Magwire MM, Bayer F, Webster CL, Cao C, Jiggins FM.
%\newblock {{S}uccessive increases in the resistance of {D}rosophila to viral
%  infection through a transposon insertion followed by a {D}uplication}.
%\newblock PLoS Genet. 2011 Oct;7(10):e1002337.

%\end{thebibliography}

\end{document}